\documentclass[11pt,letterpaper]{article}
\usepackage{emnlp2017}
\usepackage{times}
\usepackage{latexsym}

\usepackage{hyperref}
\usepackage{subfigure}
\usepackage{url}
\usepackage{amsmath}
\usepackage{mathrsfs}
\usepackage{wrapfig}
\usepackage{graphicx}
\usepackage{color}
\usepackage{appendix}
\usepackage{float}
\usepackage{outlines}
\usepackage{multirow}

\emnlpfinalcopy

\def\emnlppaperid{844}

\title{Generating High-Quality and Informative Conversation Responses \\with Sequence-to-Sequence Models}

\author{Louis Shao\textsuperscript{1}\thanks{~~Both authors contributed equally to this work.}, Stephan Gouws\textsuperscript{3}\footnotemark[1], Denny Britz\textsuperscript{2}\thanks{~~Work done as a member of the Google Brain Residency program (\texttt{g.co/brainresidency}).}, Anna Goldie\textsuperscript{2}, Brian Strope\textsuperscript{1}, Ray Kurzweil\textsuperscript{1} \\ 
{\tt \{overmind,sgouws,dennybritz,agoldie,bps,raykurzweil\}@google.com} \\
\begin{minipage}{0.45\textwidth}
\centering
\textsuperscript{1}{Google Research} \and \textsuperscript{2}{Google Brain} \\
Mountain View, CA, USA
\end{minipage}
\and
\begin{minipage}{0.45\textwidth}
\centering
\textsuperscript{3}{Google Brain \\
London, UK}
\end{minipage}
}

\date{}

\begin{document}
\maketitle

\begin{abstract}
Sequence-to-sequence models have been applied to the conversation response generation problem where the source sequence is the conversation history and the target sequence is the response. 
Unlike translation, conversation responding is inherently creative. The generation of long, informative, coherent, and diverse responses remains a hard task.
In this work, we focus on the single turn setting. 
We add self-attention to the decoder to maintain coherence in longer responses, and we propose a practical approach, called the glimpse-model, for scaling to large datasets. We introduce a stochastic beam-search algorithm with segment-by-segment reranking which lets us inject diversity earlier in the generation process.
We trained on a combined data set of over 2.3B conversation messages mined from the web. In human evaluation studies, our method produces longer responses overall, with a higher proportion rated as acceptable and excellent as length increases, compared to baseline sequence-to-sequence models with explicit length-promotion. A back-off strategy produces better responses overall, in the full spectrum of lengths.
\end{abstract}

\section{Introduction}
Building computer systems capable of general-purpose conversation is a challenging problem. However, it is a necessary step toward building intelligent agents that can interact with humans via natural language, and for eventually passing the Turing test. 
The sequence-to-sequence (\emph{seq2seq}) model has proven very popular as a purely data-driven approach in domains that can be cast as learning to map to and from variable-length sequences, with state-of-the art results in many domains, including machine translation~\citep{ChoEmnlp14,SutskeverEtc14,WuEtc16BNMT}.  Neural conversation models are the latest development in the domain of conversation modeling, with the promise of training computers to converse in an end-to-end fashion~\citep{vinyals2015neural,shang2015neural,sordoni2015neural,WenEMNLP2016}. 
Despite promising results, there are still many challenges with this approach. In particular, these models produce short, generic responses that lack diversity~\citep{sordoni2015neural,li2015diversity}. Even when longer responses are explicitly encouraged (e.g. via length normalization), they tend to be incoherent (\textit{``The sun is in the center of the sun.''}), redundant (\textit{``i like cake and cake''}), or contradictory (\textit{``I don't own a gun, but I do own a gun.''}). 

In this paper, we provide two methods to address these issues with minimal modifications to the standard seq2seq model. First, we present a \emph{glimpse model} that only trains on fixed-length segments of the target-side at a time, allowing us to scale up training to larger data sets. Second, we introduce a \emph{segment-based stochastic decoding technique} which injects diversity earlier in the generated responses. Together, we find that these two methods lead to both longer responses and higher ratings, compared to a baseline seq2seq model with explicit length and diversity-promoting heuristics integrated into the generation procedure (see Table~\ref{tab:showcase} for examples generated using our model).

In Section~\ref{sec:overview}, we present a high-level overview of these two techniques. We then discuss each technique in more detail in Sections~\ref{sec:target-attention} and \ref{sec:decode}. Finally, we report small and large-scale experimental evaluations of the proposed techniques in Section~\ref{sec:experiments}.

\section{Overview and Motivation}
\label{sec:overview}

\begin{table*}[t]
\tiny
    \centering
    \begin{tabular}{p{0.33\textwidth}|p{0.67\textwidth}}
     \hline
      \textbf{PROMPT:} & \textbf{RESPONSE FROM BASELINE (A) vs OUR METHOD (B)}\\ \hline 
      \multirow{2}{*}{\textit{what do you have planned tonight}} 
        & \textbf{A:} I have no idea what I want to do with my life. I have no idea what I want to (Bad, Mediocre, Bad) \\
        & \textbf{B:}  I was planning on doing a few of these, but I'm not sure what to choose. I need to find a place to camp, and I don't have a lot of time to make a plan. (Excellent, Good, Mediocre) \\ \hline 
        \multirow{2}{*}{\parbox{0.33\textwidth}{\textit{I'd like to read a good science fiction novel. Haven't read scifi in a while.}}} 
        & \textbf{A:}  If you like sci fi you should check out The Name of the Wind by Patrick Rothfuss (Excellent, Excellent, Excellent) \\
        & \textbf{B:} If you want to know what sci-fi is, read the first book by Neil Gaiman. It's a sci-fi sci-fi sci-fi novel. It is really good and I recommend it to all of you. It has a great premise. (Acceptable, Excellent, Acceptable) \\ \hline 
      \multirow{2}{*}{\textit{Got any big weekend plans?}} 
        & \textbf{A:} I have plans for the weekend but I don't know if I will be able to make it to the (Bad, Mediocre, Bad) \\
        & \textbf{B:} I don't really have much to do on Saturday. But I plan on going to the movies at the mall and going to the show. (Good, Good, Bad)\\ \hline       \multirow{2}{*}{\parbox{0.33\textwidth}{\textit{Sometimes flying messes my ears, because of the air pressure. Especially if I have a cold or something, flying gets me messed up.}}} 
        & \textbf{A:} I don't know why, but when I was a kid I used to do this all the time when I was younger. (Good, Bad, Bad) \\
        & \textbf{B:} I have no idea why, but I feel exactly the same way. I don't know why, but my brain is so sensitive to air that it makes my nose twitch. (Excellent, Mediocre, Good)\\ \hline
      \multirow{2}{*}{\parbox{0.33\textwidth}{\textit{Yeah. I have this friend, and we get bored just hanging out here sometimes, so we take little trips to places. We went to Seattle a couple of months ago.}}} 
        & \textbf{A:} What kind of things do you like to do? (Excellent, Excellent, Excellent) \\
        & \textbf{B:} That sounds like a great time. I have a lot of friends that go to Seattle, so I was wondering if you could share any of the locations? (Excellent, Excellent, Excellent)\\ \hline 
      \multirow{2}{*}{\textit{have you seen Suicide Squad?}} 
        & \textbf{A:} I have not, but I will check it out. (Good, Good, Good) \\
        & \textbf{B:} I saw it in a theater once and it was great. It was very... I don't know, a little dark. I like the psychological effects and the way it portrays the characters in the movie. (Excellent, Excellent, Excellent)\\ \hline 
      \multirow{2}{*}{\parbox{0.33\textwidth}{\textit{Actually I like the snow too, except for today. It's all slushy and it's all over the sidewalks and it's gross.}}} 
        & \textbf{A:} I hate the snow. I hate the snow. I hate the snow. I hate the snow. I hate the snow. I hate the snow. I hate the snow. I hate the snow. (Acceptable, Good, Good) \\
        & \textbf{B:} I'm in the same boat as you. I hate it when it's snowy and I don't like to shovel it all day. :( So this is a perfect excuse to not get out of bed and go somewhere else. (Excellent, Good, Acceptable) \\ \hline 
      \multirow{2}{*}{\textit{What do you hope to see in the future?}} 
        & \textbf{A:} I want to be able to go back in time to when I was born. I want to be (Mediocre, Bad, Mediocre) \\
        & \textbf{B:} I want to see an actual movie where the characters have the ability to hide the fact that they are actually in the movie. (Good, Acceptable, Acceptable) \\ \hline 
      \multirow{2}{*}{\textit{what dangers could happen with the singularity?}} 
        & \textbf{A:} I don't think it would be that big of a deal. (Acceptable, Mediocre, Acceptable)\\
        & \textbf{B:} If you look at it from the perspective of an alien, you can't say that there aren't any alternative ways to deal with it. (Acceptable, Mediocre, Acceptable) \\ \hline 
      \multirow{2}{*}{\parbox{0.33\textwidth}{\textit{I think you're a little crazy. }}} 
        & \textbf{A:} I don't think you know what that word means. (Acceptable, Good, Good) \\
        & \textbf{B:} I am, actually. I don't think I'm crazy though. I certainly am. (Acceptable, Mediocre, Acceptable)\\ \hline 
    \end{tabular}
    \caption{Some example responses generated from the baseline (\textbf{A}) compared to our model (\textbf{B}), with human ratings in parentheses. These were selected for quality. See the appendix for an unbiased sample.}
    \label{tab:showcase}
\end{table*}

A major difference between translation and responding to conversations is that, in the former, the high-level semantic content to generate in the target sequence $\mathbf{y}$ is completely given by the source sequence, \emph{i.e.}, given the source $\mathbf{x}$, there is low conditional entropy in the target distribution $P(\mathbf{y}|\mathbf{x})$. In the seq2seq approach, the decoder network therefore only has to keep track of where it is in the output, and the content to generate can be transformed from the relevant parts in the source via the attention mechanism~\citep{bahdanau2014neural}. In contrast, in conversation response generation, the prompt turn may be short and general (e.g., \emph{``what do you have planned tonight''}), while an appropriate response may be long and informative.

The standard seq2seq model struggles with generating long responses, since the decoder has to keep track of everything output so far in its fixed-length hidden state vector, which leads to incoherent or even contradictory outputs. To combat this, we propose to integrate \textit{target-side attention} into the decoder network, so it can keep track of what has been output so far. This frees up capacity in the hidden state for modeling the higher-level semantics required during the generation of coherent longer responses. We were able to achieve small perplexity gains using this idea on the small OpenSubtitles 2009 data set~\citep{Tiedemann09OST}. However, we found it to be too memory-intensive when scaling up to larger data sets.


As a trade-off, we propose a technique (called the `glimpse model') which interpolates between source-side-only attention on the encoder, and source and target-side attention on the encoder and decoder, respectively. Our solution simply trains the decoder on fixed-length \emph{glimpses} from the target side, while having both the source sequence and the part of the target sequence before the glimpse on the encoder, thereby sharing the attention mechanism on the encoder. This can be implemented as a simple data-preprocessing technique with an unmodified standard seq2seq implementation, and allows us to scale training to very large data sets without running into any memory issues. See Figure~\ref{fig:models-side-by-side} for a graphical overview, where we illustrate this idea with a glimpse-model of length 3.

Given such a trained model, the next challenge is how to generate long, coherent, and diverse responses with the model. As observed in the previous section and in other work, standard maximum a posteriori (MAP) decoding using beam search often yields short, uninformative, and high-frequency responses. One approach to produce longer outputs is to employ length-promoting heuristics (such as length-normalization~\citep{WuEtc16BNMT}) during decoding. We find this increases the length of the outputs, however often at the expense of coherence. Another approach to explicitly create variation in the generated responses is to rerank the $N$-best MAP-decoded list of responses from the model using diversity-promoting heuristics~\citep{li2015diversity} or a backward RNN~\cite{THWen15Stochastic}. We find this works for shorter responses, but not for long responses, primarily for two reasons: First, the method relies on the MAP-decoding to produce the $N$-best list, and as mentioned above, MAP-decoding prefers short, generic responses. Second, it is too late to delay reranking in the beam search until the whole sequence has been generated, since beam-search decoding tends to yield beams with low diversity per given prompt, even when the number of beams is high. Instead, our solution is to break up the reranking over shorter segments, and to rerank segment-by-segment, thereby injecting diversity earlier during the decoding process, where it has the most impact on the resulting diversity of the generated beams.

To further improve variation in the generated responses, we replace the deterministic MAP-decoding of the beam search procedure with sampling. If a model successfully captures the distribution of responses given targets, one can expect simple greedy sampling to produce reasonable responses. However, due to model underfitting, the learned distributions are often not sharp enough, causing step-by-step sampling to accumulate errors along the way, manifesting as incoherent outputs. 
We find that integrating sampling into the beam-search procedure yields responses that are more coherent and with more variation overall.

In summary, the contributions of this work are the following:
\begin{enumerate}
\item We propose to integrate target-side attention in neural conversation models, and provide a practical approach, referred to as the \textbf{glimpse model}, which scales well and is easy to implement on top of the standard sequence-to-sequence model.
\item We introduce a \textbf{stochastic beam-search procedure} with segment-by-segment reranking which improves the diversity of the generated responses.
\item We present large-scale experiments with human evaluations showing the proposed techniques improve over strong baselines.
\item We release our collection of context-free conversation prompts used in our evaluations as a benchmark for future open-domain conversation response research. 
\end{enumerate}

\section{Seq2Seq Model with Attention on Target}
\label{sec:target-attention}

\begin{figure*}[t]
\centering
\subfigure[The vanilla sequence-to-sequence model.]{
    \includegraphics[height=0.15\linewidth]{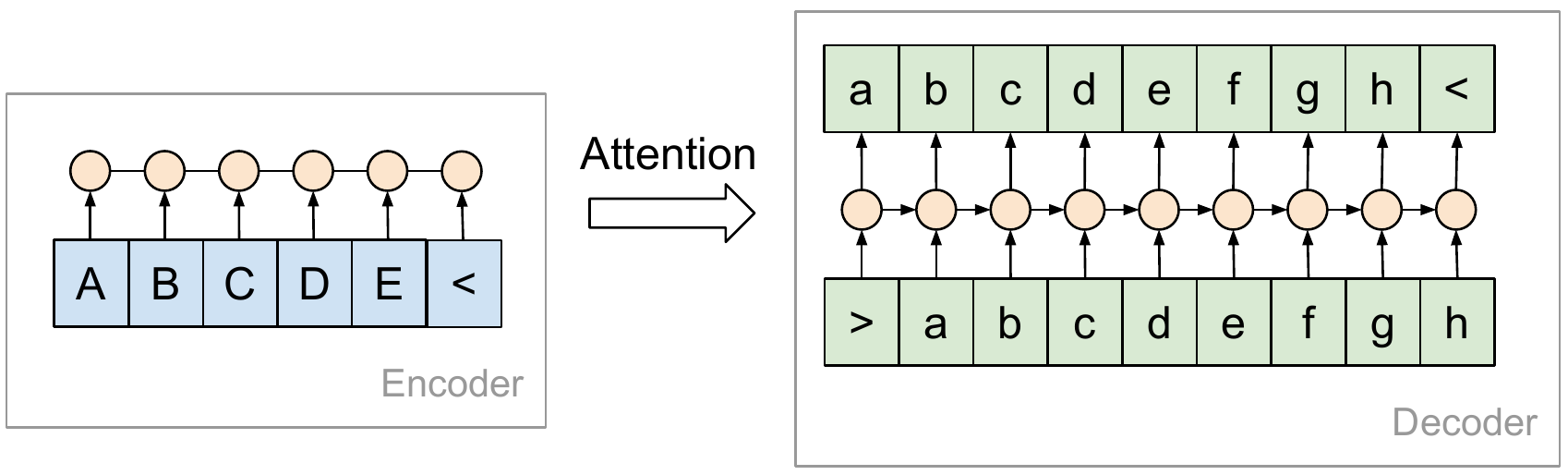}
    \label{fig:vanilla-seq2seq}
}
\subfigure[Length-3 Target-glimpse Model]{
    \includegraphics[height=0.15\linewidth]{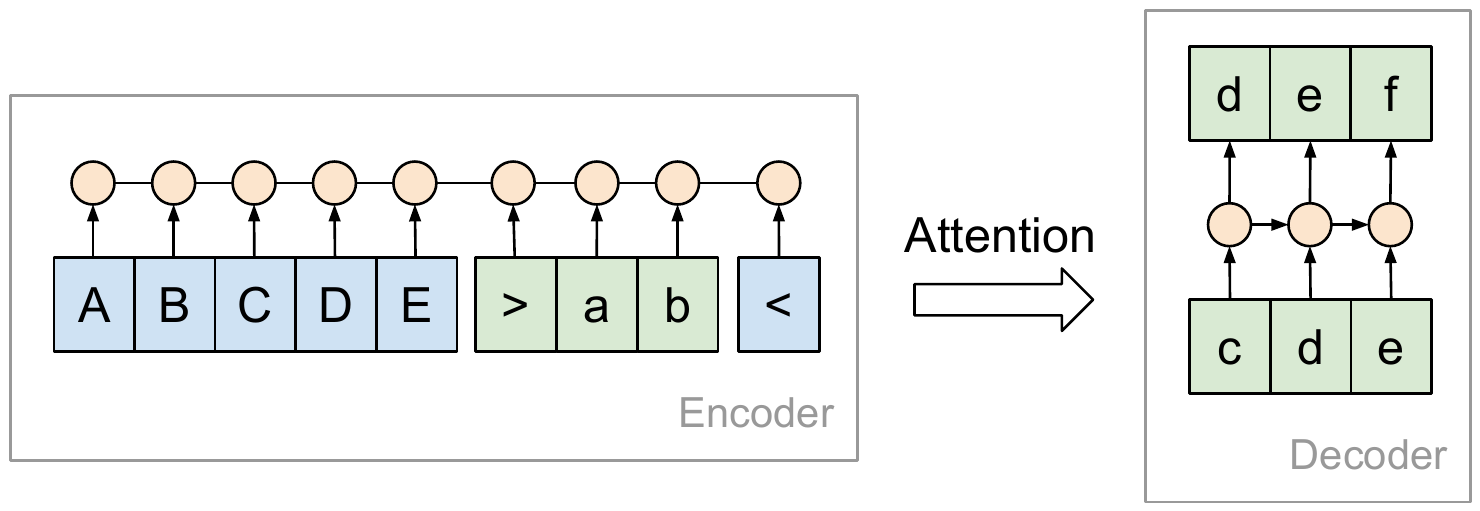}
    \label{fig:target-glimpse-model}
}
\caption{The vanilla seq2seq with attention on the left, and our proposed target-glimpse model on the right. The symbol ``\textgreater'' and ``\textless'' are start-of-sequence and end-of-sequence, respectively.}
\label{fig:models-side-by-side}
\end{figure*}

We discuss conversation response generation in the sequence-to-sequence problem setting. In this setting, there is a source sequence $\mathbf{x} = \left(x_1, x_2, ..., x_M\right)$, and a target sequence $\mathbf{y} = \left(y_0, y_1, y_2, ..., y_N\right)$. We assume $y_0$ is always the start-of-sequence token and $y_N$ is the end-of-sequence token. In a typical sequence-to-sequence model, the encoder gets its input from the source sequence $\mathbf{x}$ and the decoder models the conditional language model $P\left(\mathbf{y}|\mathbf{x}\right)$ of the target sequence $\mathbf{y}$, given $\mathbf{x}$.

Seq2seq models with attention~\citep{bahdanau2014neural} parameterize the per-symbol conditional probability as:
\begin{equation}
\label{eq-step-cond}
\begin{array}{l}
P\left( {{y_i}|{{\bf{y}}_{\left[ {0:i - 1} \right]}};{\bf{x}}} \right) = {\rm{DecoderRNN}}\left( {\color{white} \left( {{h_i}} \right)} \right.\\
{\textrm{ }\textrm{ }\textrm{ }\textrm{ }\textrm{ }\textrm{ }\textrm{ }\textrm{ }\textrm{ }\textrm{ }\textrm{ }\textrm{ }}\left. {{y_{i - 1}},{h_{i - 1}},{\rm{Attention}}\left( {{h_{i - 1}},{\bf{x}}} \right)} \right)
\end{array}
\end{equation}

for $1 \le i \le N$, where DecoderRNN() is a recurrent neural network that map the sequence of decoder symbols into fixed-length vectors, and Attention() is a function that yields a fixed-size vector summary of the encoder symbols $\mathbf{x}$ (the `focus') most relevant to predicting $y_i$, given the previous recurrent state of the network ${h}_{i-1}$ (the `context'). The full conditional probability follows from the product rule, as:
\begin{equation}
\label{eq-joint-cond}
P\left( {{\bf{y}}|{\bf{x}}} \right) = \prod\limits_{i = 1}^N {P\left( {{y_i}|{{\bf{y}}_{\left[ {0:i - 1} \right]}};{\bf{x}}} \right)} \end{equation}

We propose to implement \textbf{target-side attention} by augmenting the attention mechanism to include the part of the target sequence already generated, \emph{i.e.}, we include $\mathbf{y}_{[0:i-2]}$ in the arguments to the attention function: $\textrm{Attention}(h_{i-1}, \mathbf{y}_{[0:i-2]}, \mathbf{x})$. We implemented this in TensorFlow~\citep{tensorflow2015-whitepaper} using 3 LSTM layers on both the encoder and the decoder, with 1024 units per layer.  We experimented on the OpenSubtitles~2009 data set, and obtained a small perplexity gain from the target-side attention: 24.6 without versus 24.2 with. However, OpenSubtitles is a small data set, and the majority of its response sequences are shorter than 10 tokens. This may prevent us from seeing bigger gains, since our method is designed to help with longer outputs. In order to train on the much larger Reddit data set, we implemented this method on top of the GNMT model~\citep{WuEtc16BNMT}. Unfortunately, we met with frequent out-of-memory issues, as the 8-layer GNMT model is already very memory-intensive, and adding target-side attention made it even more so. Ideally, we would like to retain the model's capacity in order to train a rich response model, and therefore a more efficient approach is necessary. 

To this end, we propose the \textbf{target-glimpse model} which has a fixed-length decoder. The target-glimpse model is implemented as a standard sequence-to-sequence model with attention, where the decoder has a fixed length $K$. During training, we split the target sequence into non-overlapping, contiguous segments (\emph{glimpses}) with fixed length $K$, starting from the beginning. We then train on each of these glimpses, one at a time on the decoder, while putting all target-side symbols before the glimpse on the encoder. For example, if a sequence $\mathbf{y}$ is split into two glimpses $\mathbf{y}_1$ and $\mathbf{y}_2$, each with length $K$ ($\mathbf{y}_2$ may be shorter than $K$), then we will train the model with two examples, $\left(\mathbf{x} \rightarrow \mathbf{y}_1\right)$, and $\left(\mathbf{x}, \mathbf{y}_1 \rightarrow \mathbf{y}_2\right)$. Each time the concatenated sequence on the left of the arrow is put on the encoder and the sequence on the right is put on the decoder. Figure~\ref{fig:target-glimpse-model} illustrates the training of $\left(\mathbf{x}, \mathbf{y}_1 \rightarrow \mathbf{y}_2\right)$ when $K=3$. In our implementation, we always put the source-side end-of-sequence token at the end of the whole encoder sequence, and we split the glimpses according to the decoder time steps. For example, if the sequence $\mathbf{y}$ is $y_0, y_1, y_2, ..., y_{10}$, and $K=3$, the first example will have $y_0, y_1, y_2$ on the input layer of the decoder, and $y_1, y_2, y_3$ on the output layer of the decoder. The second example has $y_3, y_4, y_5$ as input of the decoder and $y_4, y_5, y_6$ as the output of the decoder, and so on. In our experiments, we use $K=10$. 

While decoding each glimpse, the decoder therefore attends to both the source sequence and the part of the target sequence that precedes the glimpse, thereby benefiting from the GNMT encoder's bidirectional RNN. Through generalization, the decoder should learn to decode a glimpse of length $K$ in any arbitrary position of the target sequence (which we will exploit in our decoding technique discussed in Section~\ref{sec:decode}). One drawback of this model, however, is that the context inputs to the attention mechanism only include the words that have been generated so far in this glimpse, rather than the words from the full target side. The workaround that we use is to simply connect the last hidden state of the GNMT-encoder to the initial hidden state of the decoder\footnote{This is the default in standard seq2seq models, but not in the GNMT model.}, thereby giving the decoder access to all previous symbols regardless of the starting position of the glimpse.

\section{Stochastic Decoding with Segment-by-Segment Reranking}
\label{sec:decode}


We now turn our attention from training to inference (decoding). Our strategy is to perform reranking with a normalized score at the segment level, where we generate the candidate segments using a trained glimpse-model and using a stochastic beam search procedure, which we discuss next. The full decoding algorithm proceeds segment by segment. 

The standard beam search algorithm generates symbols step-by-step by keeping a set of the $B$ highest-scoring beams generated so far at each step\footnote{Beams are also called `hypotheses', and $B$ is referred to as the `beam width'.}. The algorithm adds all possible single-token extensions to every existing beam, and then selects the top $B$ beams. In our stochastic beam search algorithm, we replace this deterministic top-$B$ selection by a stochastic sampling operation in order to encourage variation. Further, to discourage a single beam from dominating the search and decreasing the final response diversity, we perform a two-step sampling procedure: 1) For each single-token extension of an individual beam we don't enumerate all possibilities, but instead sample a fixed number of $D$ candidate tokens to be added to the beam. This yields a total of $B \times D$ beams, each with one additional symbol. 2) We then compute the accumulated conditional log-probabilities for each beam (normalized across all $B \times D$ beams), and treat these as the logits for sub-sampling $B$ beams for the next step. We repeat this procedure until we reach the desired segment-length $H$, or until a segment ends with the end-of-sequence token.

For a given source sequence, we can use this stochastic beam search algorithm to generate $B$ candidate $H$-length segments as the beginning of the target sequence. We then perform a reranking step (described below), and keep one of these. The concatenation of the source and the first target segment is then used as the input for generating the next $B$ candidate segments. The algorithm continues until the segment selected ends with an end-of-sequence token.

This algorithm behaves similarly to standard beam search when the categorical distribution used during the process is sharp (`peaked'), since the samples are likely to be the top categories (words) . However, when the distribution is smooth, many of the choices are likely. In conversation response generation we are dealing with a conditional probability model with high entropy, so this is what often happens in practice.

For the reranking, we normalize the scores using random prompts. In particular, suppose $\mathbf{y}_k = y_1, ..., y_{k-1}$ is a candidate segment, and $(\mathbf{x}, \mathbf{y}_{1:k-1})$ is the input to the stochastic beam search. The normalized score is then computed as follows:
\begin{equation}
S\left( {{\mathbf{y}_k} | \mathbf{x}, {\mathbf{y}_{1:k - 1}}} \right) = \frac{{P\left( {{\mathbf{y}_k} | \mathbf{x}, {\mathbf{y}_{1:k - 1}}} \right)}}{{\sum\nolimits_{\mathbf{x}' \in \Phi } {P\left( {{\mathbf{y}_k} | \mathbf{x}',{\mathbf{y}_{1:k - 1}}} \right)} }}
\end{equation}
In this equation, the set $\Phi$ is a collection of randomly sampled source sequences (prompts). In our experiments, we randomly select $Q$ prompts from the context-free evaluation set (introduced in the Experiments section).

It is worth noting that when $\Phi$ is an unbiased sample from $P(\mathbf{x})$, the summation in the denominator is a Monte-Carlo approximation of $P(\mathbf{y}_k|\mathbf{y}_{1:k-1})$. In the case of reranking whole target sequences $\mathbf{y}$, this becomes the marginal $P(\mathbf{y})$, which corresponds to the same diversity-promoting objective used in~\cite{li2015diversity}. However, we found that our approximation works better in terms of N-choose-1 accuracy (see Section~\ref{sec:exp-mot}), which suggests that its value may be closer to the true conditional probability. 

In our experiments, we set number of random prompts $Q$ to 15, segment length $H$ to 10, number of beams $B$ to 2, and samples per beam $D$ to 10. We select a small value for $B$, since we find that larger values makes the algorithm behave more like standard beam search.

\section{Experimental Results}
\label{sec:experiments}

\begin{figure*}[h]
\centering
\subfigure[]{
    \includegraphics[width=0.48\linewidth]{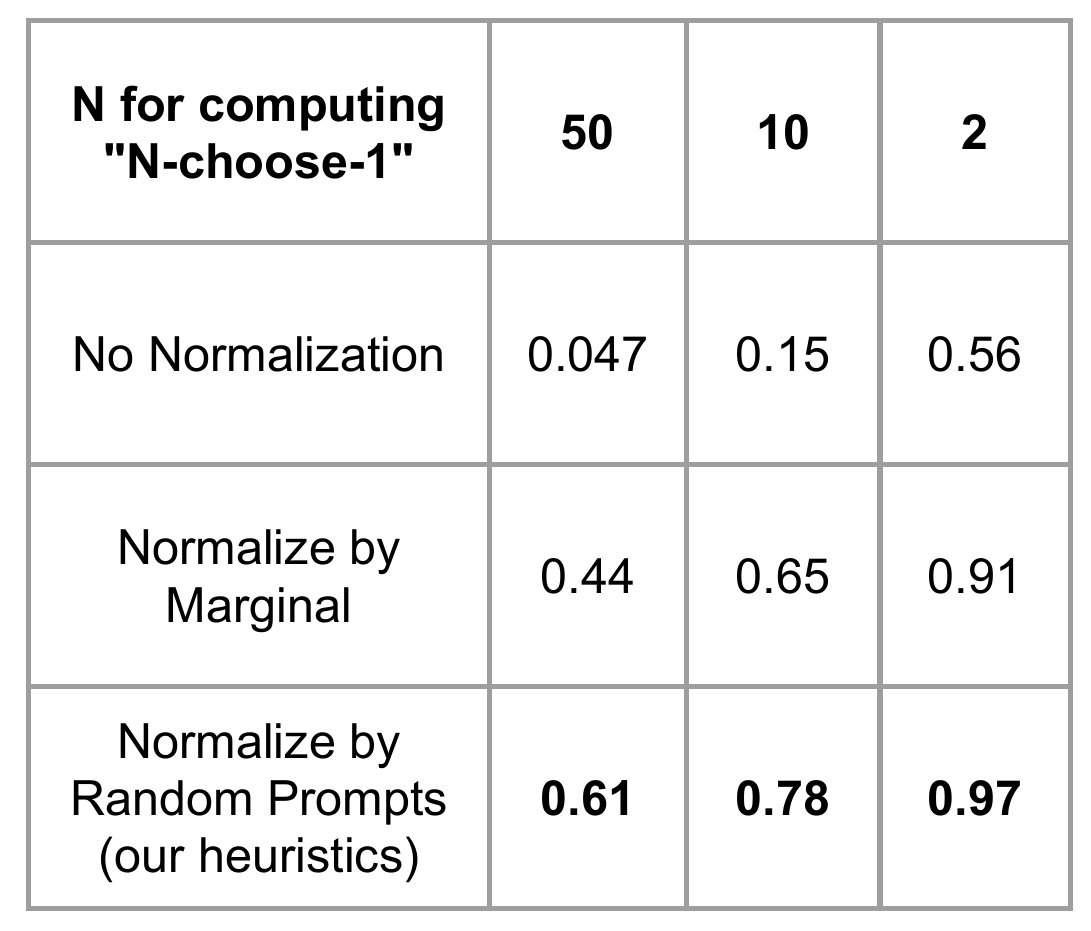}
    \label{tab:n-choose-1}
}
\subfigure[]{
    \includegraphics[width=0.48\linewidth]{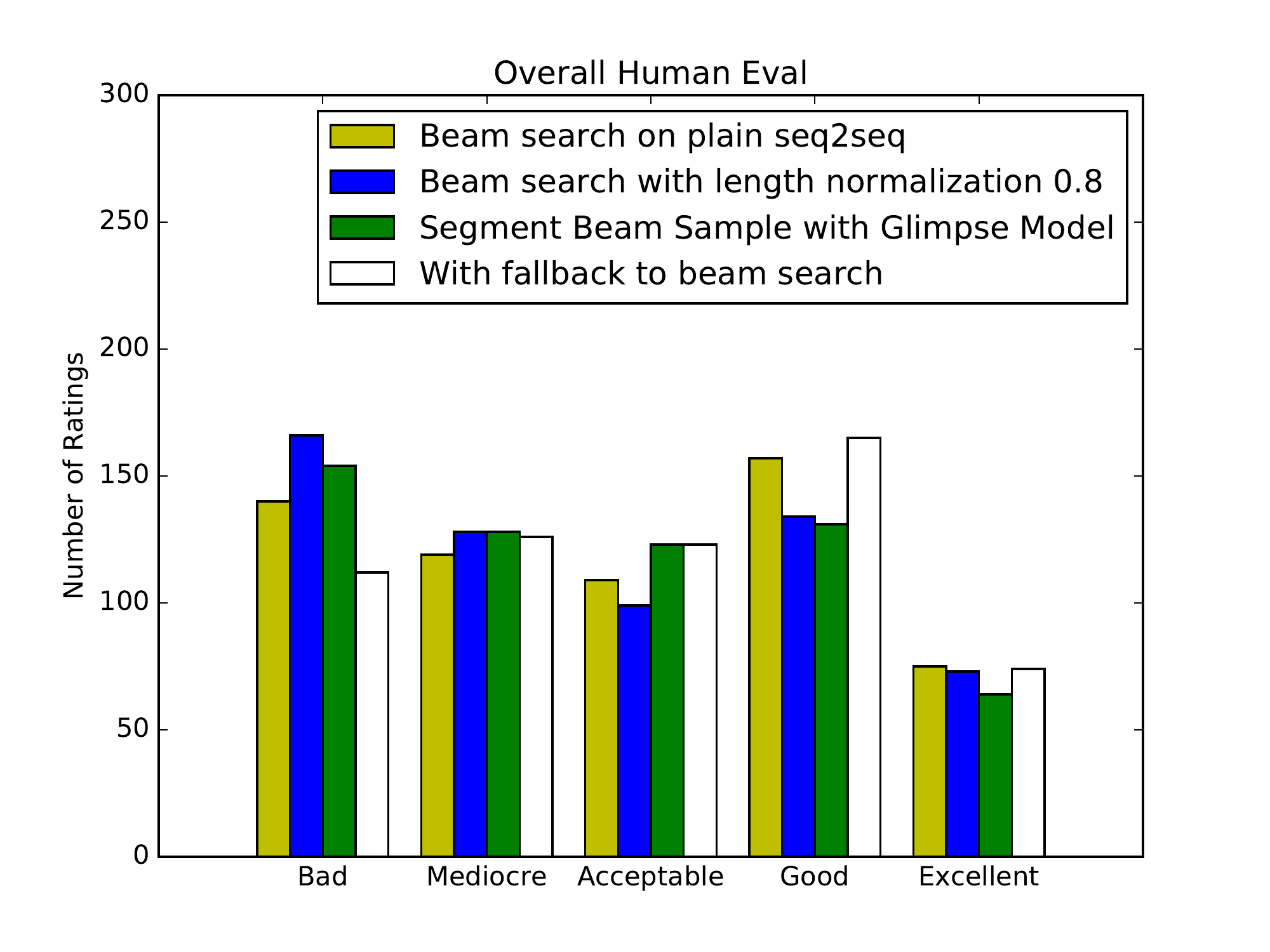}
    \label{fig:two-results-b}
}
\subfigure[]{
    \includegraphics[width=0.48\linewidth]{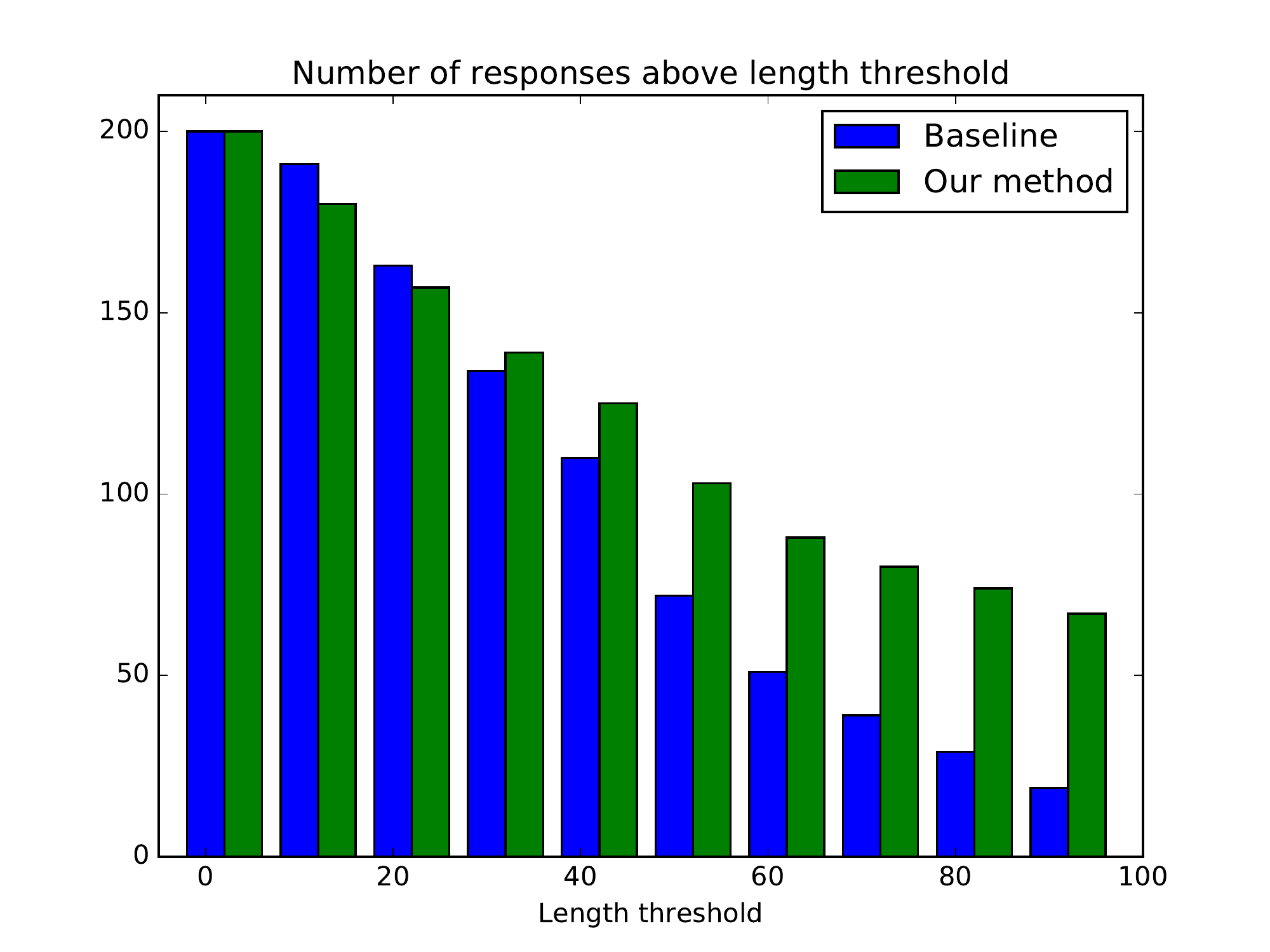}
    \label{fig:two-results-c}
}
\subfigure[]{
    \includegraphics[width=0.48\linewidth]{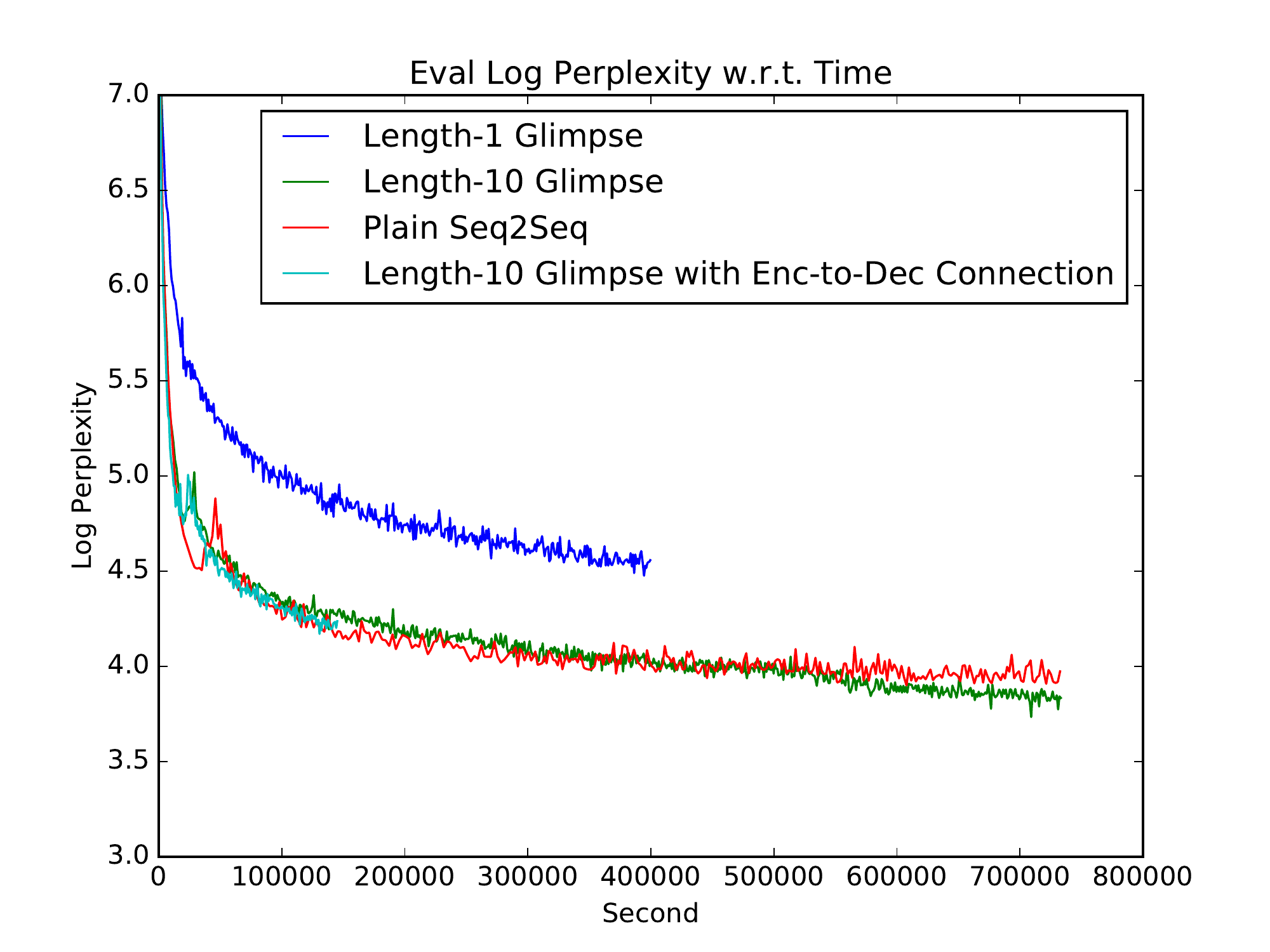}
    \label{fig-combo6-train}
}
\subfigure[]{
    \includegraphics[width=0.48\linewidth]{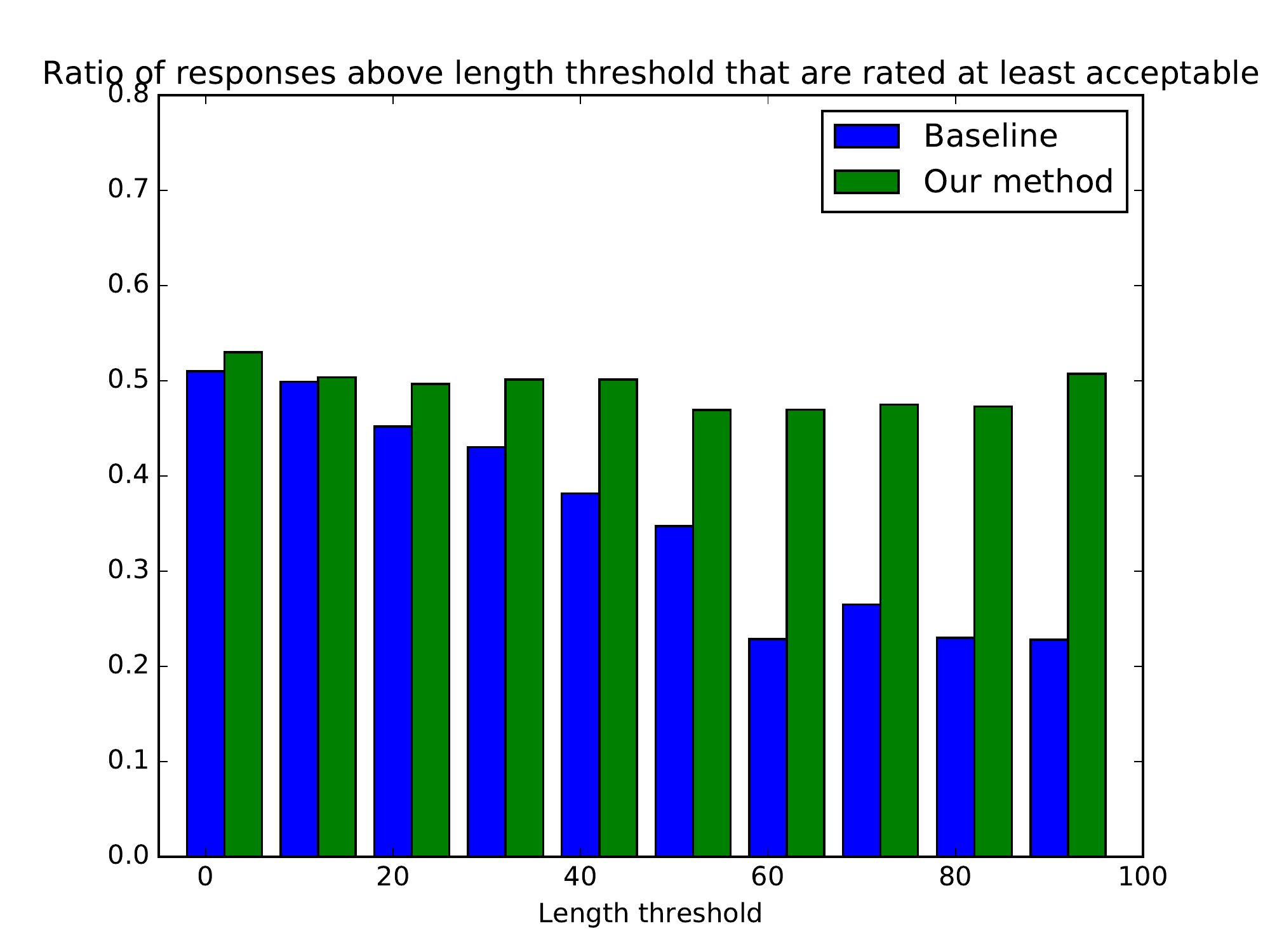}
    \label{fig:two-results-e}
}
\subfigure[]{
    \includegraphics[width=0.48\linewidth]{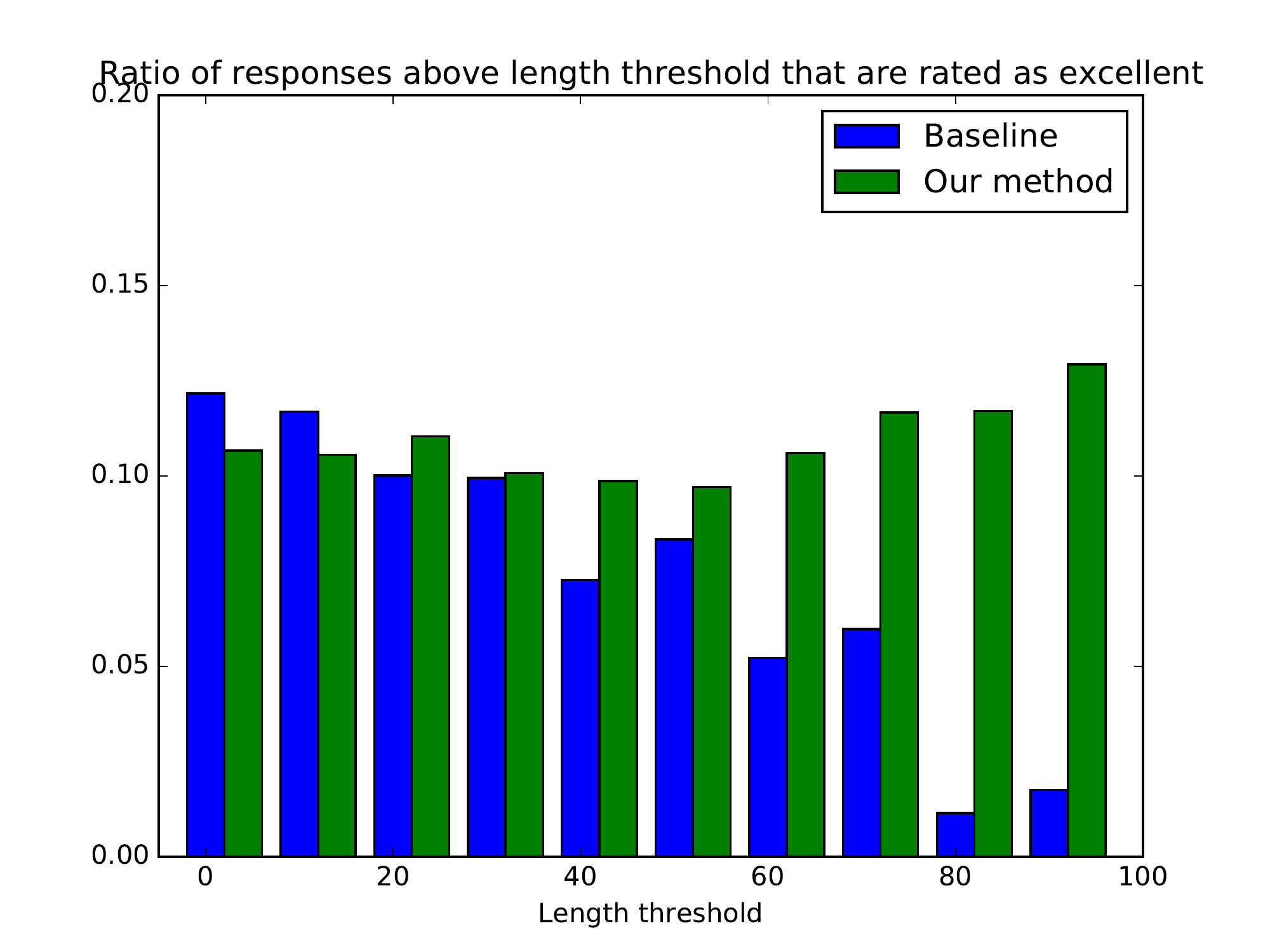}
    \label{fig:two-results-f}
}
\caption{(a) N-choose-1 evaluation on the baseline model. (d) Training progress of different models on the full combined data set. \emph{Length-1} and \emph{Length-10} are the target-glimpse models we propose, and \emph{Plain Seq2seq} is the baseline model we described. (b)(c)(e)(f): Human evaluation results on the conversation data. (b) The histogram of 5 ratings per method. (c) The length thresholds (horizontal axis) and the number of responses generated that are above the length threshold (vertical axis); (e) The proportion of responses above the length-threshold that are judged at least \emph{Acceptable}; (f) The proportion of responses above the length-threshold that are judged as \emph{Excellent}. The length thresholds are all measured in number of characters.}
\label{fig:two-results}
\end{figure*}

In this section we present experimental results for evaluating the target-glimpse model and the stochastic decoding method that we presented. We train the model using the Google neural machine translation model (GNMT, \cite{WuEtc16BNMT}), on a data set that combines multiple sources mined from the Web:

\begin{enumerate}
\item The full Reddit data\footnote{Download links are at https://redd.it/3bxlg7} that contains 1.7 billion messages (221 million conversations).
\item The 2009 Open Subtitles data (0.5 million conversations, \cite{Tiedemann09OST}).
\item The Stack Exchange data (0.8 million conversations).
\item Dialogue-like texts that we recognized and extracted from the web (17 million conversations).
\end{enumerate}

For all these data sets, we extract pairs of messages where one can be considered as a response to the other. For example, in the Reddit data set, the messages belonging to the same post are organized as a tree. A child node is a message that replies to its parent. This may not necessarily be true as people may be replying to other messages that are also visually close. However, for our current single-turn experiments, we treat these as a single exchange.

In this setting, the GNMT model trained on prompt-to-response pairs works surprisingly well without modification when generating short responses with beam search. Similar to previous work on neural conversation models, we find that the generated responses are almost always grammatical, and sometimes even interesting. They are also usually on topic. In addition, we found that even greedy sampling from the 8-layer GNMT model produces grammatical responses most of the time, although these responses are more likely to be semantically-broken than responses generated using standard beam search. We would like to leverage the benefits of greedy sampling, because the induced variation generates more surprises and may potentially help improve user-engagement, and we found that our proposed segment-based beam sampling procedure accomplishes this to some extent.

\subsection{Evaluation Metric}

It is difficult to come up with an objective evaluation metric for conversation response generation that can be computed automatically. The conditional distribution $P\left(\mathbf{y}|\mathbf{x}\right)$ is supposed to have high entropy in order to be interesting (many possible valid responses to a given prompt). Therefore BLEU scores used in translation are not a good fit (also see~\cite{LiuEMNLP2016}). Other than looking at the evaluation set perplexity, we use two metrics, the \textbf{N-choose-1 accuracy} and \textbf{5-scale side-by-side human evaluation}. In the N-choose-K metric, we use the model as a retriever. Given a prompt, we ask the model to rank $N$ candidate responses, where one is the ground truth and the other $N-1$ are random responses from the same data set. We then calculate the N-choose-K accuracy as the proportion of trials where the true response is in the top $K$. The prompts used for evaluation are selected randomly from the same data set. This metric isn't necessarily correlated well with the true response quality, but provides a useful first diagnostic for faster experimental iteration. It takes about a day to train a small model on a single GPU that reaches 2-choose-1 accuracies of around 70\% or 80\%, but it is much harder to make progress on the 50-choose-1 accuracy. As a reference, human performance on the 10-choose-1 task is around 45\% accuracy.

In the 5-scale human evaluation, we use a collection of 200 context-free prompts\footnote{This list will be released to the community.}. These prompts are collected from the following sources, and filtered to prompts that are context-free (\emph{i.e.} do not depend on previous turns in the conversation), general enough, and by eliminating near duplicates:
\begin{enumerate}
\item The questions and statements that users asked an internal testing bot.
\item The Fisher corpus~\citep{David04thefisher}.
\item User inputs to the Jabberwacky chatbot\footnote{http://www.jabberwacky.com/}.
\end{enumerate}

These can be either generic or specific.  Some example prompts from this collection are shown in Table~\ref{tab:showcase}. These prompts are open-domain (not about any specific topic), and include a wide range of topics. Many require some creativity for answering, such as \emph{``Tell me a story about a bear.''} Our evaluation set is therefore not from the same distribution as our training set. However, since our goal is to produce good general conversation responses, we found it to be a good general purpose evaluation set.

The evaluation itself is done by human raters. They are well-trained for the purpose of ensuring rating quality, and they are native English speakers. The A 5-scale rating is produced for each prompt-response pair: \emph{Excellent}, \emph{Good}, \emph{Acceptable}, \emph{Mediocre}, and \emph{Bad}. For example, the instructions for rating \emph{Excellent} is ``On topic, interesting, shows understanding, moves the conversation forward. It answers the question.'' The instruction for \emph{Acceptable} is ``On topic but with flaws that make it seem like it didn’t come from a human. It implies an answer.'' The instruction for \emph{Bad} is ``A completely off-topic statement or question, nonsensical, or grammatically broken.  It does not provide an answer.''

In our experiments, we perform the evaluations side-by-side, each time using responses generated from two methods. Every prompt-response pair is rated by three raters. We rate 200 pairs in total for every method, garnering 600 ratings overall. After the evaluation, we report aggregated results from each method individually.

\subsection{Motivating Experiments}
\label{sec:exp-mot}

To see whether generating long responses is indeed a challenging problem, we trained the plain seq2seq with the GNMT model where the encoder holds the source sequence and the decoder holds the target sequence. We experimented with the standard beam search and the beam search with length normalization $\alpha=0.8$ similar to~\cite{WuEtc16BNMT}. With this length normalization the generated responses are indeed longer. However, they are more often semantically incoherent. It produces \emph{``I have no idea what you are talking about.''} more often, similarly observed in \cite{Li16DeepRe}. The human evaluation results are summarized in Figure~\ref{fig:two-results-b}. Methods that generate longer responses have more \emph{Bad} and less \emph{Excellent} / \emph{Good} ratings. 

We also performed the N-choose-1 evaluation on the baseline model using different normalization schemes. The results are shown in Table~\ref{tab:n-choose-1}. \emph{No Normalization} means that we use $P(\mathbf{y}|\mathbf{x})$ for scoring, \emph{Normalize by Marginal} uses $P(\mathbf{y}|\mathbf{x}) / P(\mathbf{y})$, as suggested in~\cite{li2015diversity}, and \emph{Normalize by Random Prompts} is our scoring objective described in Section~\ref{sec:decode}. The significant boost when using both normalization schemes indicates that the conditional log probability predicted by the model may be biased towards the language model probability of $P(\mathbf{y})$. After adding the normalization, the score may be closer to the true conditional log probability.


Overall, this \emph{reranking} evaluation indicates that our heuristic is preferred to scoring using the marginal. However, it is unfortunately hard to directly make use of this score during beam search decoding (\emph{i.e., generation}), since the resulting sequences are usually ungrammatical, as also observed by~\cite{li2015diversity}. This is the motivation for using a segment-by-segment reranking procedure, as described in Section~\ref{sec:decode}.

\subsection{Large-Scale Experiments}

For our large-scale experiments, we train our target-glimpse model on the full combined data set. Figure~\ref{fig-combo6-train} shows the training progress curve. In this figure, we also include the curve for $K=1$, that is, the glimpse model with decoder-length 1. It is clear enough that this model progresses much slower, so we terminated it early. However, it is surprising that the glimpse model with $K=10$ progresses faster than the baseline model with only source-side attention, because the model is trained on examples with decoder-length fixed at 10, while the average response length is 38 in our data set. This means it takes on average 3.8x training steps for the glimpse model to train on the same number of raw training-pairs as the baseline model. Despite this, the faster progress indicates that target-side attention indeed helps the model generalize better.


The human evaluation results shown in Figure~\ref{fig:two-results} compare our proposed method with the baseline seq2seq model. For this, we trained a length-10 target-glimpse model and decoded with stochastic beam-search using segment-by-segment reranking. In our experiments, we were unable to generate better long, coherent responses using the whole-sequence level reranking method from \cite{li2015diversity} compared to using standard beam search with length-normalization\footnote{This is because the method reranks the responses in the $N$-best list resulting from the beam search, which tend to be short with not much variation to begin with.}. We therefore choose the latter as our baseline, because it is the only method which generates responses that are long enough that we can compare to.

Figure~\ref{fig:two-results} shows that our proposed method generates more long responses overall. One third of all responses are longer than 100 characters, while the baseline model produces only a negligible fraction. Although we do not employ any length-promoting objectives in our method, length-normalization is used for the baseline. For responses generated by our method, the proportion of \emph{Acceptable} and \emph{Excellent} responses remains constant or even increases as the responses grow longer. Conversely, human ratings decline sharply with length for the baseline model.

The percentage of test cases with major agreement is high for both methods. We consider a test to have major agreement if two ratings out of the three are the same. For the baseline method, 80\% of the responses have major agreements, and for our method it is 70\%.

However, shorter responses have a much smaller search space, and we find that standard beam search tends to generate better (``safer'') short responses. To maximize cumulative response quality, we therefore implemented a back-off strategy that combines the strengths of the two methods. We choose to fallback to the baseline model without length normalization when the latter produces a response shorter than 40 characters, otherwise we use the response from our method. This corresponds to the white histogram in Figure~\ref{fig:two-results-b}. Compared to the other methods in the figure, the combined strategy results in more ratings of \emph{Excellent}, \emph{Good}, \emph{Acceptable}, and \emph{Mediocre}, and fewer \emph{Bad} ratings. With this strategy, among the responses generated for the same 200 prompts, 133 were from the standard beam search and 67 were from our model. Out of the 67 long responses, two thirds were longer than 60 characters and half were longer than 75 characters. To compare the combined model's performance with the baseline, we generated responses from both models using the same 200 prompts. For 20 of the response pairs, human raters had no preference, but for the remaining 180, human raters preferred the combined model's response in 103 cases and the baseline's in only 77, indicating a significant win.

\section{Conclusion}

The research of building end-to-end systems that can engage in general-purpose conversation is still in its infancy. More significant progress is expected to be made with more advanced neural architectures. However, our results reported in this paper show that minimal modeling change and a slightly more advanced decoding technique, combined with training over very large data sets, can still lead to noticeable improvements in the quality of responses generated using neural conversation models. 
Overall, we found using fixed-lengths in the decoder to make it easier to train on large data sets, as well as to allow us to improve the diversity and coherence of the generated responses earlier during generation, when it has most impact.  
While the focus of this work has been on conversation modeling, we expect some of these results to carry over to other sequence-to-sequence settings, such as machine translation or image-captioning.

\subsubsection*{Acknowledgments}

We would like to thank Quoc Le, Oriol Vinyals and Jakob Uszkoreit for many helpful discussions, and Scott Benson, Fuchun Peng for collecting the context-free prompt set, and Amin Ahmad for setting up the human evaluation, and Rami Eid, Daniel Cer for collecting training data sets, and Yonghui Wu, Zhifeng Chen, Mike Schuster for help on training the GNMT model.

\bibliography{references}
\bibliographystyle{emnlp_natbib}

\appendix

\section{The Context-Free Prompt List and Evaluation Results}

The following are samples from each rating scale, for both our method (without the fallback strategy) and the baseline method. For each rating we show 5 randomly selected examples. These were selected purely at random, and not cherry-picked, in order to show an unbiased sample of the types of responses the methods generate.

\subsection{Our Method}

\begin{outline}[enumerate]
\scriptsize
\1 Excellent
\2 Prompt: \textit{what does your handwriting look like?} \\
Response: \textit{I have a pen on the left hand side.}
\2 Prompt: \textit{have you seen Suicide Squad?} \\
Response: \textit{I saw it in a theater once and it was great. It was very... I don't know, a little dark. I like the psychological effects and the way it portrays the characters in the movie.}
\2 Prompt: \textit{I do not own an Iphone} \\
Response: \textit{I own an iphone}
\2 Prompt: \textit{what are you watching right now?} \\
Response: \textit{watch the new season of rick and morty}
\2 Prompt: \textit{Actually I like the snow too, except for today. It's all slushy and it's all over the sidewalks and it's gross.} \\
Response: \textit{I'm in the same boat as you. I hate it when it's snowy and I don't like to shovel it all day. :( So this is a perfect excuse to not get out of bed and go somewhere else.}
\1 Good
\2 Prompt: \textit{I went on a bike ride today.} \\
Response: \textit{That is awesome.}
\2 Prompt: \textit{how was your weekend? what'd you do?} \\
Response: \textit{I had a good weekend and I had a good time. I just wanted to get some rest.}
\2 Prompt: \textit{get anything good for xmas?} \\
Response: \textit{I got a bunch of stuff to give away, and I don't know how long it will take to send. I know I have a lot of stuff that I want to give to people and I don't know if I can give it away to someone else.}
\2 Prompt: \textit{What do you want to be?} \\
Response: \textit{I want to be a psychologist.}
\2 Prompt: \textit{Nothing makes me happy. :(.} \\
Response: \textit{I can't even imagine how awful this feels to be.}
\1 Acceptable
\2 Prompt: \textit{Are you good at math?} \\
Response: \textit{I have to admit, I am a terrible coder...}
\2 Prompt: \textit{I didn't used to like Indian food, but I think at this point I like just about anything.} \\
Response: \textit{Indian indian food is so good.}
\2 Prompt: \textit{i was in bed all day yesterday} \\
Response: \textit{how do you feel?}
\2 Prompt: \textit{It sounds like you're a sports fan.} \\
Response: \textit{I am, I watch a lot of baseball and I don't want to be a sports fan.}
\2 Prompt: \textit{I've never learned how to fly} \\
Response: \textit{You have no idea how to fly a plane. It's not that hard.}
\1 Mediocre
\2 Prompt: \textit{I live alone, and if I eat at home I eat whatever I can put in the microwave that night. A lot of times it's just that, or I'll trade with friends; I'll cook and they'll cook and then we'll swap leftovers.} \\
Response: \textit{I live alone, and if I eat at home, I eat whatever I can put in the microwave that night. I do this all the time. I eat the same thing over and over again.}
\2 Prompt: \textit{I don't eat seafood. I don't eat anything that used to swim. I know that's kind of rigid, but I had a bad experience when I was a kid. Just doesn't do it for me.} \\
Response: \textit{I don't eat anything that used to swim. That's not the point. The point is that you can eat fish that will eat whatever you want. That's just not the same thing as fish.}
\2 Prompt: \textit{I find I like Japanese food a lot. I can make a number of the types of rolls and various types of sushi. Of course I end up doing a lot of things with noodle soup just because it's relatively cheap and fast.} \\
Response: \textit{I don't understand, how would you go about getting a sushi roll with Japanese food?}
\2 Prompt: \textit{Let's talk about cooking. Do you own a steamer?} \\
Response: \textit{I own a steamer but my user agent is a hunter, so he owns a steamer, but I don't own one, he uses it a lot but I used to own a steamer.}
\2 Prompt: \textit{Have you heard of Turing?} \\
Response: \textit{I think I have. I like it.}
\1 Bad
\2 Prompt: \textit{I've had a cold myself this week. Right now I'm trying to figure out how to get rid of it.} \\
Response: \textit{I have a cold and I don't have any colds because I have a cold.}
\2 Prompt: \textit{How much time would you say you spend attached to a computer? 2 hours? 10 hours?} \\
Response: \textit{I have a gaming computer that I'd rather not have to spend hours on, so I spend all day playing games on it on a regular basis.}
\2 Prompt: \textit{How significant is the potential for nuclear energy?} \\
Response: \textit{I think there is a very good chance it will be a very strong nuclear weapon, but it wouldn't make a huge impact. The only thing that has a chance of making it into nuclear power is a nuclear disaster, which will be very very very unlikely if it ever does.}
\2 Prompt: \textit{I really like the trees of san francisco} \\
Response: \textit{I like the new one too}
\2 Prompt: \textit{I don't eat seafood. I don't eat anything that used to swim. I know that's kind of rigid, but I had a bad experience when I was a kid. Just doesn't do it for me.} \\
Response: \textit{I don't eat anything that used to swim. That's not the point. The point is that you can eat fish that will eat whatever you want. That's just not the same thing as fish.}
\end{outline}

\subsection{Baseline Method}

\begin{outline}[enumerate]
\scriptsize
\1 Excellent
\2 Prompt: \textit{What do you think of Asimov's novels?} \\
Response: \textit{I read them when I was in high school and I loved them.}
\2 Prompt: \textit{Finally we're getting a winter with some snow.} \\
Response: \textit{Where is this? I want to go to there.}
\2 Prompt: \textit{Are you a bot?} \\
Response: \textit{I am a bot, and this action was performed automatically.}
\2 Prompt: \textit{Are you at work now?} \\
Response: \textit{Yes, I am.}
\2 Prompt: \textit{Actually, we just got a snow storm yesterday. I live in Maryland. Where do you live, on the east coast?} \\
Response: \textit{I live on the east coast of the US.}
\1 Good
\2 Prompt: \textit{Do you read the bible?} \\
Response: \textit{I don't read the bible. I read the bible. I read the bible. I read the bible}
\2 Prompt: \textit{have you seen Suicide Squad?} \\
Response: \textit{I have not, but I will check it out.}
\2 Prompt: \textit{Are you at work now?} \\
Response: \textit{Yes, I am.}
\2 Prompt: \textit{I run alot. That's my jam. I run a marathon a year.} \\
Response: \textit{Do you run on a treadmill or a treadmill?}
\2 Prompt: \textit{how was your dentist appt?} \\
Response: \textit{It was a lot better than I thought it would be.}
\1 Acceptable
\2 Prompt: \textit{What do you think of Asimov's novels?} \\
Response: \textit{I read them when I was in high school and I loved them.}
\2 Prompt: \textit{is chicken healthier than beef?} \\
Response: \textit{Yes, it is.}
\2 Prompt: \textit{It's a beautiful day.} \\
Response: \textit{I wish I could upvote this more than once.}
\2 Prompt: \textit{Let's talk about cooking. Do you own a steamer?} \\
Response: \textit{I have a steamer and a steamer.}
\2 Prompt: \textit{what r u doing this summer?} \\
Response: \textit{I have no idea what I want to do with my life.}
\1 Mediocre
\2 Prompt: \textit{What do you think about universal healthcare?} \\
Response: \textit{I don't like the idea of universal healthcare. I don't like the idea of universal healthcare. I think}
\2 Prompt: \textit{It will be very cool when we succeed in sending astronauts to Mars.} \\
Response: \textit{I think it would be a good idea to send astronauts to Mars.}
\2 Prompt: \textit{I've a software engineer at Google, it's a good life} \\
Response: \textit{I have no idea what you are talking about.}
\2 Prompt: \textit{I taught in Kansas. I taught French and Spanish at high school and college level.} \\
Response: \textit{Did you learn it in high school?}
\2 Prompt: \textit{What makes you happy?} \\
Response: \textit{The fact that I don't have to worry about my}
\1 Bad
\2 Prompt: \textit{i need to start eating healthier} \\
Response: \textit{Your submission has been automatically removed pending manual approval because your account is either too new or does not meet}
\2 Prompt: \textit{where did you grow up?} \\
Response: \textit{where are you from?}
\2 Prompt: \textit{i hate being the center of attention} \\
Response: \textit{I hate it when people do this.}
\2 Prompt: \textit{You can't eat pasta every day. You can't live on only Italian food.} \\
Response: \textit{I don't eat Italian food, but I do eat a lot of Italian food.}
\2 Prompt: \textit{You are a program, and I am human.} \\
Response: \textit{I am a bot, and this action was performed automatically. Please contact the moderators of this subreddit if you have}
\end{outline}

\end{document}